\documentclass[letterpaper, 10 pt, conference]{ieeeconf}  

\IEEEoverridecommandlockouts                              

\usepackage{url}
\usepackage{caption}
\usepackage{subcaption}
\usepackage{graphicx}
\usepackage{cite}
\usepackage{multirow}
\usepackage{enumerate}
\usepackage{amsmath}
\usepackage{etoolbox}
\usepackage{placeins}
\usepackage{hyperref}
\usepackage{setspace} 
\usepackage{romannum}
\usepackage{wasysym}

\usepackage{bbding}
\usepackage{pifont}
\usepackage{xcolor}
\usepackage[outline]{contour}
\usepackage{booktabs}
\usepackage{amsmath, amssymb, amsfonts}
\usepackage{bm} 

\usepackage{makecell} 

\newcommand{\real}{\mathbb{R}}
\newcommand{\calD}{\mathcal{D}}

\newcommand{\vect}[1]{\mathbf{#1}} 

\DeclareMathAlphabet\mathbfcal{OMS}{cmsy}{b}{n}

\title{\LARGE \bf 3D Flow Diffusion Policy: \\ Visuomotor Policy Learning via Generating Flow in 3D Space}

\author{Sangjun Noh$^{*}$, Dongwoo Nam$^{*}$, Kangmin Kim, Geonhyup Lee, Yeonguk Yu, Raeyoung Kang, Kyoobin Lee†
\thanks{\text{*} Equally contributed.}
\thanks{All authors are with the School of Integrated Technology (SIT), Gwangju Institute of Science and Technology (GIST), Cheomdan-gwagiro 123, Buk-gu, Gwangju 61005, Republic of Korea. 
† Corresponding author: Kyoobin Lee {\tt\small kyoobinlee@gist.ac.kr}}%
}

\begin{document}

\maketitle
\thispagestyle{empty}
\pagestyle{empty}

\begin{abstract}
Learning robust visuomotor policies that generalize across diverse objects and interaction dynamics remains a central challenge in robotic manipulation. Most existing approaches rely on direct observation-to-action mappings or compress perceptual inputs into global or object-centric features, which often overlook localized motion cues critical for precise and contact-rich manipulation. We present \underline{\textbf{3D}} \underline{\textbf{F}}low \underline{\textbf{D}}iffusion \underline{\textbf{P}}olicy (\textbf{3D FDP}), a novel framework that leverages scene-level 3D flow as a structured intermediate representation to capture fine-grained local motion cues. Our approach predicts the temporal trajectories of sampled query points and conditions action generation on these interaction-aware flows, implemented jointly within a unified diffusion architecture. This design grounds manipulation in localized dynamics while enabling the policy to reason about broader scene-level consequences of actions. Extensive experiments on the MetaWorld benchmark show that 3D FDP achieves state-of-the-art performance across 50 tasks, particularly excelling on medium and hard settings. Beyond simulation, we validate our method on eight real-robot tasks, where it consistently outperforms prior baselines in contact-rich and non-prehensile scenarios. These results highlight 3D flow as a powerful structural prior for learning generalizable visuomotor policies, supporting the development of more robust and versatile robotic manipulation. Robot demonstrations, additional results, and code can be found at \href{https://sites.google.com/view/3dfdp/home}{https://sites.google.com/view/3d-fdp}.

\end{abstract}

\section{Introduction}

Learning robust manipulation skills in unstructured environments is a fundamental challenge in robotics. Policies deployed in such settings must generalize across diverse object geometries, appearances, dynamics, and contextual variations. Visual imitation learning offers a scalable alternative by enabling robots to acquire visuomotor skills from expert demonstrations without requiring task-specific reward functions~\cite{jang2022bc, cliport2021}. Recent approaches~\cite{brohan2022rt, team2024octo, kim2024openvla, black2410pi0, bjorck2025gr00t} often employ end-to-end architectures that map perception directly to control, leveraging transformers whose attention mechanisms excel at capturing global context and integrating multiple modalities.

More recently, diffusion models have shown significant promise in this domain~\cite{chi2023diffusion, team2024octo}. By formulating action generation as an iterative denoising process, they provide stable training and support complex, multimodal action distributions. When combined with structured inputs such as point clouds, these models achieve strong performance across diverse manipulation tasks~\cite{ze20243d, ke20243d, gervet2023act3d}. Despite these advances, most existing methods still follow a direct observation-to-action mapping, often compressing perceptual inputs into global or object-centric features. Such representations can overlook localized motion cues that are essential for precise and contact-rich manipulation.  While some works introduce intermediate representations, approaches based on object poses~\cite{su2025motion, xian2023chaineddiffuser} can struggle to capture scene-level dynamics, and those predicting future frames~\cite{black2023zero, du2023learning, liang2025video, ni2024generate, luo2024grounding, zhao2025cot} are often computationally intensive.

\begin{figure}[!t]
    \centering
        \begin{subfigure}[t]{\columnwidth}
            \centering
            \includegraphics[width=\textwidth]{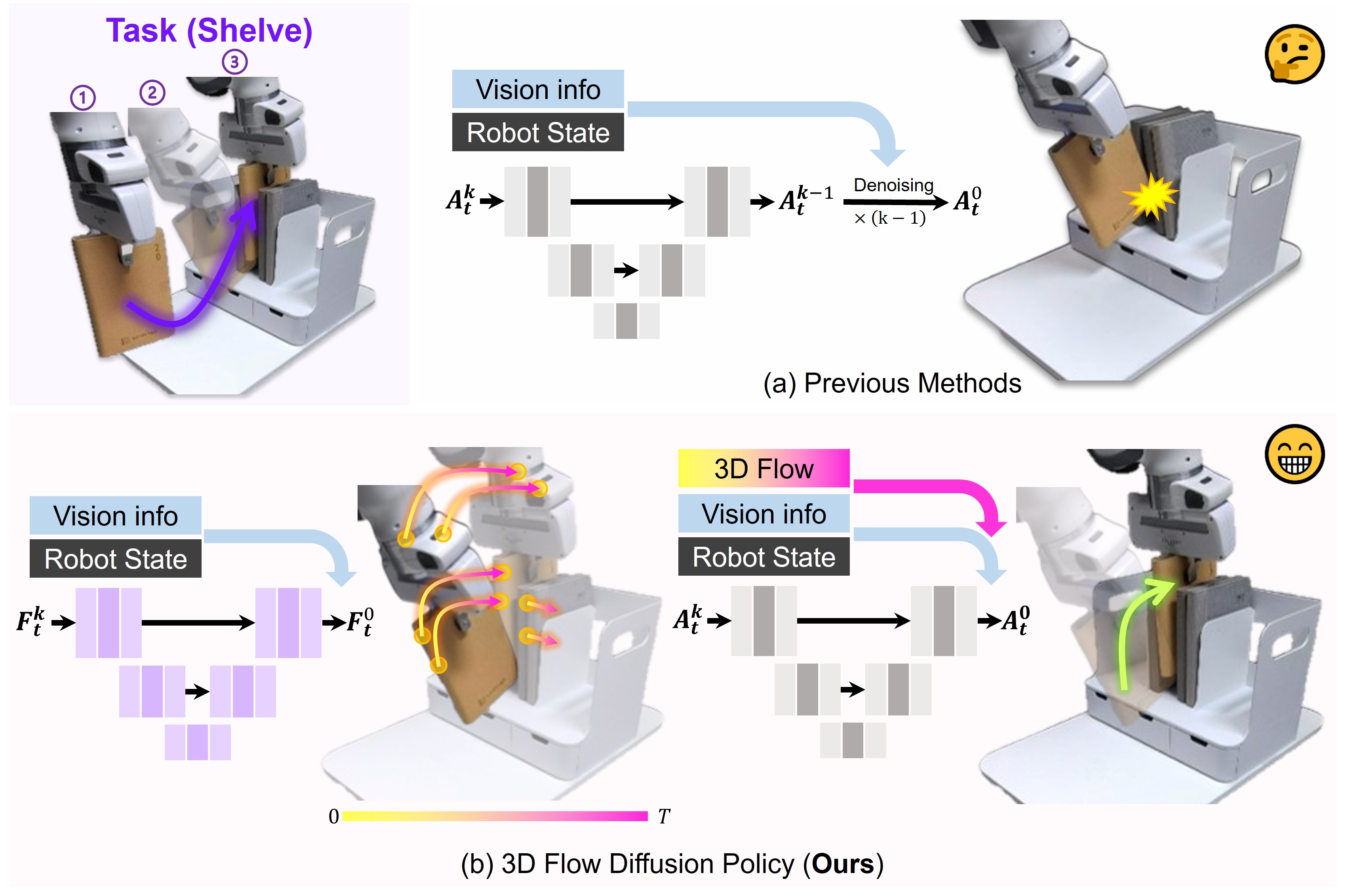}
        \end{subfigure}
        
    \caption{Comparison between prior diffusion policy methods and our proposed 3D FDP. Previous approaches predict actions from raw observations (vision and robot state), whereas 3D FDP first predicts 3D flows as an intermediate representation and then generates actions conditioned on both the observations and the predicted flows.}
    \label{fig:concept}
\end{figure}

To address these limitations, we introduce the \underline{\textbf{3D}} \underline{\textbf{F}}low \underline{\textbf{D}}iffusion \underline{\textbf{P}}olicy (\textbf{3D FDP}), an architecture that explicitly models local interaction dynamics through an intermediate representation of 3D scene flow. Our method predicts the temporal trajectories of sampled 3D query points and uses this predicted flow to infer actions. Unlike prior approaches~\cite{su2025motion, yang2025gripper} that rely on object-level representations, our model learns both flow prediction and action generation jointly within a unified diffusion framework (Fig. \ref{fig:concept}). This integrated design offers two key advantages. First, 3D flow captures fine-grained correlations between the robot gripper and manipulated objects, providing spatial grounding that supports contact-aware behavior. Second, by modeling how local motion propagates across the scene, the policy can reason about the downstream effects of manipulation, such as how inserting a book might disturb surrounding objects on a cluttered shelf. These capabilities improve generalization across both geometric and dynamic variations.

We evaluate 3D FDP on the MetaWorld benchmark~\cite{metaworld2019} and a set of real-world manipulation tasks. On MetaWorld, our approach achieves state-of-the-art performance across all difficulty levels, with particularly strong improvements on medium and hard tasks. In real-robot experiments involving diverse physical interactions, 3D FDP demonstrates capabilities beyond baseline policies, successfully handling contact-rich and non-prehensile tasks where previous methods fail. These results demonstrate that 3D scene flow provides an effective inductive bias for learning generalizable visuomotor policies. Our main contributions are as follows:
\begin{itemize}
    \item We introduce a scene-level 3D flow representation that captures interaction-aware motion throughout the scene.
    \item We develop a unified diffusion-based architecture that jointly learns 3D flow prediction and action generation in an end-to-end manner.
    \item We demonstrate that 3D FDP achieves state-of-the-art results on 50 MetaWorld tasks and outperforms prior baselines in real-world manipulation scenarios.
    \item We provide a comprehensive analysis that demonstrates the effectiveness of our 3D flow representation, investigating key aspects such as conditioning strategies and query point density.
\end{itemize}

\section{Related Works}
\subsection{Visual Imitation Learning}
Visual imitation learning has become a central paradigm for acquiring robotic manipulation skills, enabling policies to learn directly from expert demonstrations without manually engineered reward functions. Early works such as BC-Z~\cite{jang2022bc} and RISE~\cite{wang2024rise} demonstrated that visuomotor policies could be trained from demonstration data but often exhibited limited generalization. These models typically relied on flat behavior cloning architectures with minimal inductive biases or temporal reasoning.

Recent advances have introduced stronger structural priors into policy design. A significant development is the use of diffusion models to generate actions as a denoising process. Diffusion Policy~\cite{chi2023diffusion} provides a framework for stable training and multi-modal action synthesis, while 3D Diffusion Policy~\cite{ze20243d} extends this concept to 3D spatial representations. Concurrently, approaches like Motion Before Action~\cite{su2025motion} condition policies on object motion predicted from prior observations. Hierarchical diffusion variants further enhance policy structure by decomposing long-horizon trajectories into intermediate keypoints or subgoals~\cite{xian2023chaineddiffuser, ma2024hierarchical, wang2025hierarchical}.

Vision-language-action (VLA) models~\cite{brohan2022rt, team2024octo, kim2024openvla, black2410pi0, bjorck2025gr00t}, including RT-1, Octo, and OpenVLA, have expanded policy learning to multimodal settings by conditioning on textual goals or instructions. While these models demonstrate impressive scalability, their reliance on global feature representations can limit their performance on contact-rich tasks requiring fine-grained precision. Despite the growing diversity in model architectures, most approaches continue to rely on direct observation-to-action mappings, which can abstract away the local motion patterns crucial for manipulation. To address these limitations, our work leverages scene-level 3D flow as a structured intermediate representation that grounds policy learning in localized interaction dynamics.

\subsection{Flow-Based Representations for Robotic Manipulation}
Flow is a foundational representation in computer vision, where optical flow~\cite{huang2022flowformer, shi2023flowformer++} and point tracking~\cite{karaev2024cotracker, karaev24cotracker3, xiao2024spatialtracker} are used to estimate motion, establish temporal correspondence, and reconstruct object trajectories. These methods provide dense temporal information that informs downstream tasks such as segmentation, tracking, and 3D reconstruction.

In robotics, early flow-based approaches often supported manipulation through heuristic object tracking or manual pose alignment, which restricted their adaptability. More recent work leverages learned 2D flow for policy conditioning. Track2Act~\cite{bharadhwaj2024track2act} refines heuristic trajectories with residual learning; ATM~\cite{wen2023any} conditions policies on pixel-level motion for articulated or deformable objects; and Im2Flow2Act~\cite{xu2024flow} extracts transferable flow from third-person videos to enable one-shot policy transfer. However, these models operate in the image space, lacking the 3D spatial precision essential for manipulation tasks.

To address this, several methods incorporate 3D information. For example, recent works have begun to use 3D flow as a basis for world models~\cite{zhi20253dflowaction} or as a foundational representation of affordances for various tasks~\cite{yuan2024general}. Concurrently, PPI~\cite{yang2025gripper} significantly improves spatial grounding by introducing object-level 3D point flow as a policy input. While highly effective, these approaches are often object-centric, as their flow estimation operates specifically on the surfaces of predefined objects. Furthermore, in many existing frameworks, flow is treated as an auxiliary input; that is, the action generation process is not explicitly structured to be consistent with the predicted motion. In contrast, our method establishes scene-level, query-point 3D flow as a core, learnable intermediate representation. Instead of merely being an input, this dynamic flow is predicted first and then directly informs the subsequent generation of the robot's action within a unified framework.

\begin{figure*}[ht!]
    \centering
        \includegraphics[width=\textwidth]{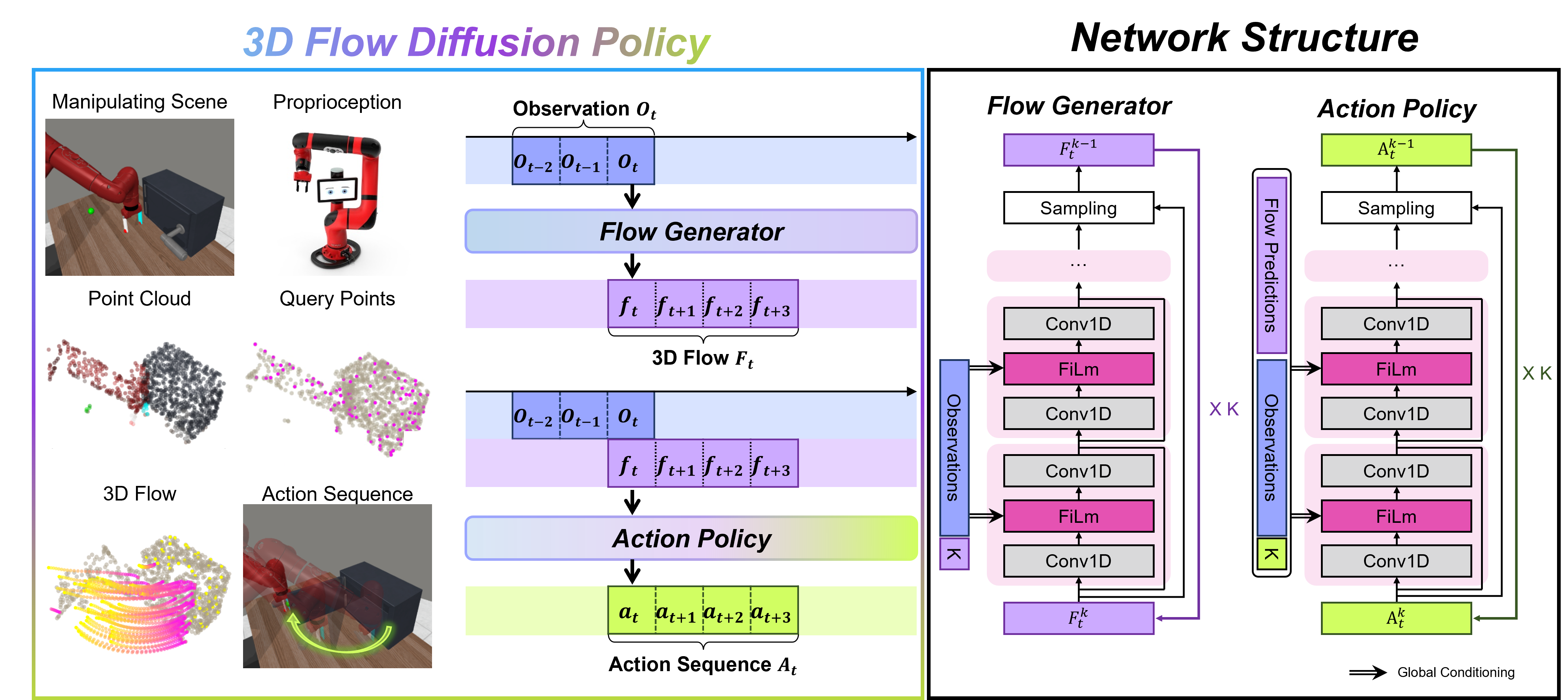}
    \caption{\textbf{Overview of 3D Flow Diffusion Policy.} (Left) The policy takes observations as input, including point clouds, sampled query points, and proprioception, and first predicts query-point trajectories with the flow generator. These predicted flows then condition the action policy to generate deployable robot actions. (Right) Detailed architecture of the flow generator and action policy modules.}

    \label{fig:graspsam}
\end{figure*}

\section{Problem Statement}

Our objective is to learn a visuomotor policy that generates low-level robot actions by explicitly reasoning about local interaction dynamics, represented as scene-level 3D flow over sampled query points.

\noindent\textbf{Observations and Actions.}
At each timestep $t \in \{1, \ldots, T\}$, the robot receives an observation $\vect{o}_t = (\vect{p}_t, \vect{s}_t)$, where:
\begin{itemize}
    \item $\vect{p}_t \in \real^{N_p \times 3}$ is a point cloud captured from a single fixed RGB-D camera with known intrinsics and extrinsics.
    \item $\vect{s}_t \in \real^{d_s}$ denotes the robot's proprioceptive state (e.g., joint angles or end-effector pose).
\end{itemize}
The robot executes a continuous low-level action $\vect{a}_t \in \real^{d_a}$, such as an end-effector displacement or pose command. A history of the past $T_h$ observations is denoted as $\vect{O}_{t-T_h:t} = (\vect{o}_{t-T_h}, \dots, \vect{o}_t)$. A demonstration trajectory of horizon $T$ is defined as $\bm{\tau} = \{(\vect{o}_1, \vect{a}_1), \ldots, (\vect{o}_T, \vect{a}_T)\}$, and a dataset of $N$ demonstrations is denoted by $\calD = \{\bm{\tau}_i\}_{i=1}^{N}$.

\noindent\textbf{Query Points and 3D Scene Flow.} 
To represent interaction dynamics, we sample $M$ query points $\{ \vect{q}_j \in \real^3 \}_{j=1}^M$ from the initial point cloud $\vect{p}_1$. The 3D flow for a single query point $\vect{q}_j$ at timestep $t$ is defined as its displacement from its initial position:
\[
    \Delta \vect{q}_{j,t} = \vect{q}_{j,t} - \vect{q}_{j,t-1},
\]
where $\vect{q}_{j,t}$ is the 3D position of the $j$-th query point at time $t$. 
The full 3D flow representation for the entire trajectory is the set of all such displacement trajectories:
\[
    \vect{F}_{t:t+T_f} = ( \{\Delta \vect{q}_{j,t}\}_{j=1}^M, \dots, \{\Delta \vect{q}_{j,t+T_f-1}\}_{j=1}^M ).
\]
\noindent We denote a future sequence of flow displacements over a horizon $T_f$ starting from time $t$ as $\vect{F}_{t:t+T_f}$.

\noindent\textbf{Policy Formulation.}
Instead of learning a direct observation-to-action mapping~\cite{chi2023diffusion, ze20243d, team2024octo}, our goal is to learn a policy $f_\theta$ that maps a history of observations $\vect{O}_{t-T_h:t}$ to a future sequence of actions $\hat{\vect{A}}_{t:t+T_a}$. We decompose this policy into two learnable components: a flow predictor $g_\theta$ and an action generator $h_\theta$. First, the flow predictor generates a future flow sequence $\hat{\vect{F}}_{t:t+T_f}$ based on the observation history:
\[
    g_\theta: \vect{O}_{t-T_h:t} \rightarrow \hat{\vect{F}}_{t:t+T_f}.
\]
Then, the action generator produces the final action sequence prediction conditioned on both the observation history and the predicted flow:
\[
    h_\theta: (\vect{O}_{t-T_h:t}, \hat{\vect{F}}_{t:t+T_f}) \rightarrow \hat{\vect{A}}_{t:t+T_a}.
\]
The full policy is a composition of these two modules, $f_\theta = h_\theta \circ g_\theta$. This formulation allows the policy to leverage localized motion cues in 3D space as a structured intermediate for robust action generation.

\section{Method}
\subsection{Motivation}
Inspired by the efficiency of modern visual trackers (e.g., CoTracker~\cite{karaev2024cotracker, karaev24cotracker3}, SpatialTracker~\cite{xiao2024spatialtracker}) and the successes of flow-based robotic manipulation~\cite{bharadhwaj2024track2act, wen2023any, xu2024flow}, we address the object-centric focus that has been a key limitation of prior work~\cite{su2025motion, yang2025gripper}. We argue that achieving complex manipulation requires a scene-level understanding of dynamics. Therefore, our method adopts 3D flow not as a tool for tracking a single object, but as a holistic representation of the entire scene's motion, captured through a sparse set of query points. This shift from an object-centric to a scene-centric perspective enables our policy to reason about broader interaction contexts, unlocking more challenging manipulation capabilities.

\subsection{Perception}
\noindent\textbf{3D Scene Representation.} We represent the manipulation scene as a 3D point cloud generated from a single-view RGB-D camera, converting depth images into 3D coordinates using known camera parameters. We crop out static background regions and sample a fixed number of points using Farthest Point Sampling (FPS) to ensure spatial coverage. While some methods like PPI~\cite{yang2025gripper} incorporate semantic features from large pretrained models (e.g., DINOv2~\cite{oquab2023dinov2}), we rely solely on geometric information to reduce computational cost and facilitate real-world deployment. The point cloud is encoded using a modified PointNet~\cite{qi2017pointnet} architecture proposed in DP3~\cite{ze20243d}, extended to output both a global scene feature and point-wise local features.

\noindent\textbf{Query Point Sampling.} We sample a set of $M$ query points from the first frame of the observation sequence to serve as anchors for flow prediction. Farthest Point Sampling (FPS) is used to select spatially distributed points that cover both the object and its surrounding context. These points are tracked throughout the episode, allowing the model to learn interaction-aware motion conditioned on the initial scene configuration.

\subsection{3D Scene Flow}
We obtain ground-truth scene flow trajectories using different methods for simulation and real-world settings. In MetaWorld, we can leverage the simulator's access to ground-truth mesh poses. We sample query points from the initial depth-based point cloud and associate each point with the nearest vertex on the object mesh. We then reconstruct each query point’s 3D trajectory by transforming its associated vertex using the known mesh poses over time.

In real-world environments, obtaining ground-truth motion is more challenging. Prior works often rely on motion capture systems (MBA~\cite{su2025motion}) or complex object reconstruction pipelines (e.g., BundleSDF~\cite{wen2023bundlesdf}, FoundationPose~\cite{wen2024foundationpose}), which can be impractical as they require expensive hardware or intensive post-processing. Instead, we generate 3D flow trajectories through a more scalable process. First, we sample a set of 3D query points from a point cloud, derived from an initial single-view depth image, using Farthest Point Sampling (FPS). These 3D points are then projected onto the corresponding RGB frame to define their initial 2D pixel locations. We then employ CoTracker~\cite{karaev2024cotracker, karaev24cotracker3} to predict the 2D trajectories of these points across the episode. To enhance robustness against potential tracking failures from occlusions in long sequences, we perform inference in overlapping temporal chunks (e.g., processing frames 0-32, then 1-33, and so on) rather than on the entire sequence at once. This strategy effectively minimizes the accumulation of tracking errors. Finally, the resulting 2D trajectories are lifted back into 3D space using the corresponding depth information at each timestep, yielding continuous 3D flow trajectories in a lightweight and efficient manner.


\subsection{Flow-Conditioned Policy Learning}

Our policy is composed of two conditional diffusion models that implement the flow predictor $g_\theta$ and the action generator $h_\theta$ defined in the Problem Statement.

\noindent\textbf{3D Flow Generation.}
The flow predictor $g_\theta$ is implemented as a denoising diffusion model. Its goal is to generate the predicted flow sequence $\hat{\vect{F}}_{t:t+T_f}$ by progressively refining a noise tensor. To guide this process, we first compute a vector $\hat{\vect{O}}^{\text{flow}}_t$ from the observation history $\vect{O}_{t-T_h:t}$. This is achieved by using both the global scene feature and the per-point local features from the PointNet encoder. For each query point, we process its local feature with an MLP and concatenate the result with the global scene feature, producing a set of vectors that are projected to form the final condition representation, denoted as $\hat{\vect{O}}^{\text{flow}}_t$. The core of this module is a denoising network, denoted as $\epsilon_\theta^{\text{flow}}$, which takes a noisy flow sequence $\vect{F}_k \sim \mathcal{N}(0, I)$ and the condition vector as input. At each diffusion step $k$, it predicts the noise to be removed:
\begin{equation}
    \vect{F}_{k-1} = \alpha_k \vect{F}_k - \gamma_k \, \epsilon_\theta^{\text{flow}}(\vect{F}_k, \hat{\vect{O}}^{\text{flow}}_t, k) + \sigma_k \mathcal{N}(0, I),
\end{equation}
where $\{\alpha_k, \gamma_k, \sigma_k\}$ follow a pre-defined noise schedule.

\noindent\textbf{Action Generation.}
Similarly, the action generator $h_\theta$ is implemented as a conditional diffusion model that produces the final action sequence $\hat{\vect{A}}_{t:t+T_a}$. This module is conditioned on both the observation history and, crucially, the predicted flow sequence $\hat{\vect{F}}_{t:t+T_f}$ generated by the first module. To form its condition vector $\hat{\vect{O}}^{\text{act}}_t$, we first encode the predicted flow sequence $\hat{\vect{F}}_{t:t+T_f}$ using a 1D temporal convolution followed by global average pooling to produce a plan-level embedding. This embedding is then concatenated with the global observation feature. The resulting vector $\hat{\vect{O}}^{\text{act}}_t$ remains fixed across all denoising steps. The corresponding denoising network, $\epsilon_\theta^{\text{act}}$, refines a noisy action sequence $\vect{A}_k \sim \mathcal{N}(0, I)$ to generate the final output:
\begin{equation}
    \vect{A}_{k-1} = \alpha_k \vect{A}_k - \gamma_k \, \epsilon_\theta^{\text{act}}(\vect{A}_k, \hat{\vect{O}}^{\text{act}}_t, k) + \sigma_k \mathcal{N}(0, I).
\end{equation}
This hierarchical structure allows the policy to align action generation with both the high-level observation context and the flow-derived summary of future interaction dynamics.

\subsection{Implementation Details} 
\noindent\textbf{Training.}
We jointly train the flow predictor $g_\theta$ and the action generator $h_\theta$ using two L1 objectives: the flow prediction loss $\mathcal{L}_F$ and the action prediction loss $\mathcal{L}_A$.
\begin{align}
    \mathcal{L}_F &= \|\epsilon_\theta^{\text{flow}}(\vect{F}_k, \hat{\vect{O}}^{\text{flow}}_t, k) - \vect{F}\|_1, \\
    \mathcal{L}_A &= \|\epsilon_\theta^{\text{act}}(\vect{A}_k, \hat{\vect{O}}^{\text{act}}_t, k) - \vect{A}\|_1,
\end{align}
where $\vect{F}$ and $\vect{A}$ are the ground-truth sequences. We use separate AdamW optimizers for each module with a learning rate of $1\times10^{-4}$, betas of $(0.95, 0.999)$, and weight decay of $1\times10^{-6}$. The learning rate follows a cosine decay schedule with $500$ warmup steps. While parameters for the PointNet encoder are shared, the diffusion models for flow and action are separate. We use a DDIM~\cite{song2020denoising} scheduler with $100$ training timesteps. We train $1000$ epochs with a batch size of $128$. Each input point cloud consists of $N_p=1024$ points. We use an observation history of $T_h=2$ and predict an action horizon of $T_f=T_a=16$. We apply exponential moving average (EMA) to the model parameters and use the EMA model for evaluation. Training for each model was conducted on a single NVIDIA RTX 3090 GPU and took approximately 6 hours to complete.

\noindent\textbf{Inference.}
For inference, we use a consistent procedure across all experiments. We sample M=100 query points once from the initial observation at the start of an episode. At each timestep, the policy sequentially predicts the future flow, which then conditions the generation of a 16-step action sequence produced by a DDIM scheduler with 10 inference steps. From this sequence, we use an 8-step action horizon and execute the first action.

\begin{figure}[t!]
    \centering
        \begin{subfigure}[t]{\columnwidth}
            \centering
            \includegraphics[width=0.9\textwidth]{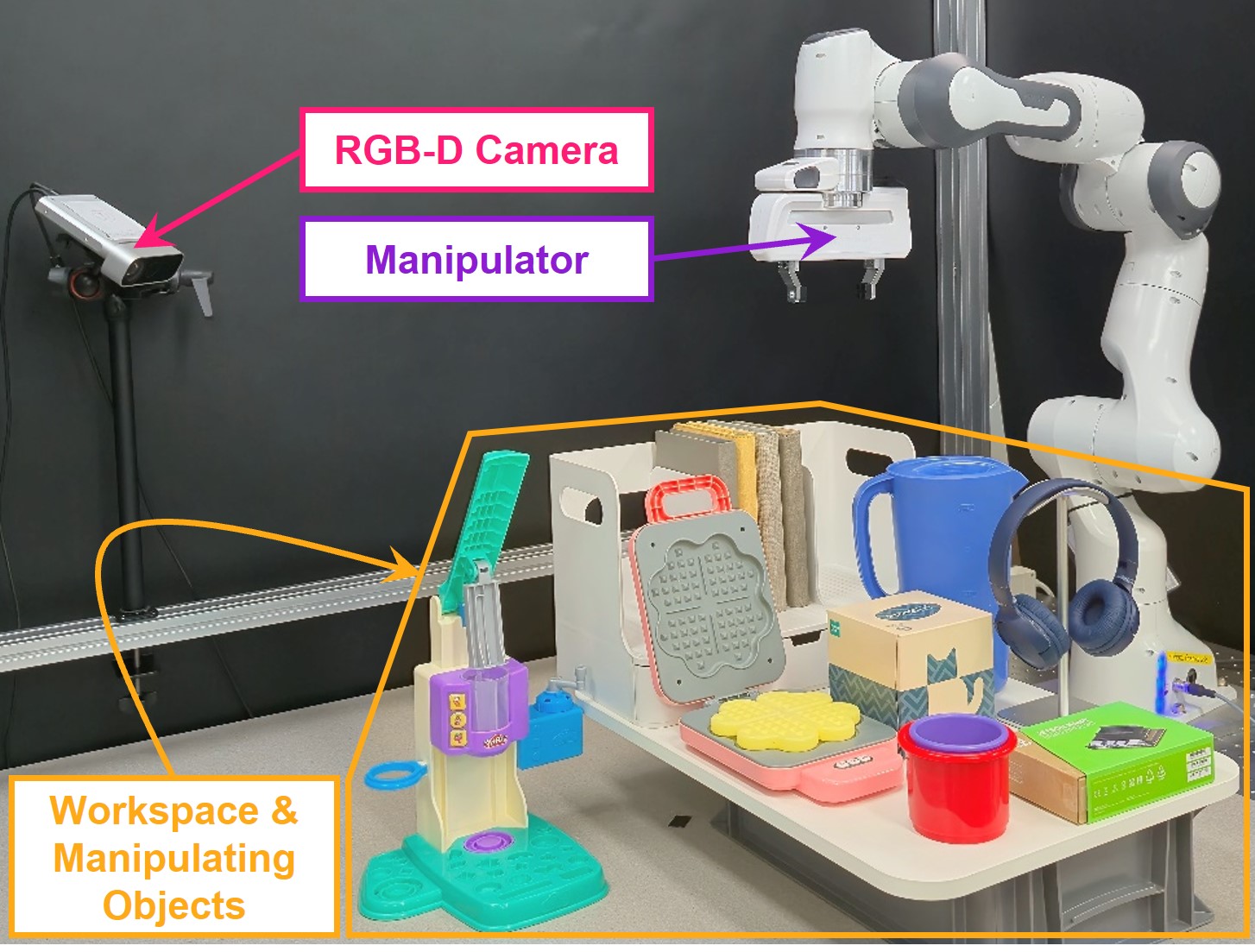}
        \end{subfigure}
        
    \caption{\textbf{Our Real-World Experimental Setup.} The setup consists of a Franka Panda manipulator and a fixed single RGB-D camera. This configuration is used to collect expert demonstrations and evaluate policy performance on our 8 real-world benchmark tasks.}
    \label{fig:realworld_setup}
\end{figure}
\begin{table}[ht]
\caption{\textbf{MetaWorld Results.} Mean success rates (\%) on 50 tasks, categorized by difficulty. 3D FDP outperforms both baselines, DP3~\cite{ze20243d} and MBA~\cite{su2025motion}, across all levels.}
\label{tab:main_table}
\resizebox{\columnwidth}{!}{%
\begin{tabular}{@{}c|c|c|c|c@{}}
\toprule
Method & \begin{tabular}[c]{@{}c@{}}MetaWorld \\ Easy (28)\end{tabular} & \begin{tabular}[c]{@{}c@{}}Metaworld\\ Medium (11)\end{tabular} & \begin{tabular}[c]{@{}c@{}}Metaworld\\ Hard (6)\end{tabular} & \begin{tabular}[c]{@{}c@{}}Metaworld\\ Very Hard (5)\end{tabular} \\ \midrule
DP3~\cite{ze20243d}    & 85.8  & 51.1   & 52.4    & 46.5         \\
MBA*~\cite{su2025motion}   & 84.2  & 49.7   & 53.0    & 46.9         \\
3D FDP (Ours) & \textbf{86.5}  & \textbf{55.3}   & \textbf{55.6}    & \textbf{47.9}         \\ \bottomrule
\end{tabular}%
}
\end{table}

\section{Experiments}
\subsection{Simulation Experiments}

\noindent\textbf{Environment and Dataset.}
We conduct simulation experiments in MetaWorld~\cite{metaworld2019}, a benchmark featuring 50 diverse manipulation tasks. To ensure a fair comparison, all methods are trained on a consistent dataset consisting of 10 expert demonstrations per task. Our approach is supervised with 3D scene flow, while the MBA baseline is supervised with 9-DoF object poses, both using the same 3D point clouds as input. This design isolates the performance difference to the choice of intermediate representation.

\noindent\textbf{Baselines.}
We compare our method against two baselines representing distinct approaches to visuomotor control:
\begin{itemize}
\item \textbf{DP3}~\cite{ze20243d}: A strong baseline that learns a direct observation-to-action mapping without an explicit intermediate representation.
\item \textbf{MBA}~\cite{su2025motion}: A state-of-the-art method that utilizes predicted 9-DoF object poses as its intermediate representation.
\end{itemize}
For the MBA baseline, we replace the flow prediction head in our model with a pose prediction head, keeping all other components unchanged to strictly isolate the contribution of the intermediate representation.

\noindent\textbf{Evaluation Protocol.}
To ensure robust and reproducible results, policies are trained once (seed 0) and evaluated across three different seeds (0, 1, 2). The primary metric is the mean success rate, consistent with prior work~\cite{ze20243d, su2025motion}. For the main comparative results, this rate is averaged over 200 evaluation episodes per seed (600 total episodes per task). This large number of episodes is necessary to mitigate the performance variance commonly observed in diffusion-based policies. For supplementary experiments, results are averaged over 20 episodes per seed.

\begin{table}[]
\caption{\textbf{Effect of Query Points.} Performance on the MetaWorld hard set with different numbers of query points (NQ). Tracking more query points consistently improves success rates by providing richer interaction cues.}
\centering

\label{tab:effect_of_nq}
\resizebox{0.85\columnwidth}{!}{%
\begin{tabular}{@{}cccccccl@{}}
\toprule
                         & \multicolumn{7}{c}{Metaworld Hard}                                      \\ \midrule
\multicolumn{1}{c|}{NQ}  & A     & HI   & POH  & PP    & PB    & \multicolumn{1}{c|}{P}& Avg. \\ \midrule
\multicolumn{1}{c|}{50}  & 83.3          & 3.3          & 26.7 & 46.7 & 96.7 & \multicolumn{1}{c|}{56.7} & 52.2     \\
\multicolumn{1}{c|}{100} & \textbf{93.3} & 5.0          & 31.7 & 58.3 & 98.3 & \multicolumn{1}{c|}{61.7} & 58.1     \\
\multicolumn{1}{c|}{200} & 88.3          & \textbf{8.3} & \textbf{43.3} & \textbf{60.0} & \textbf{98.3} & \multicolumn{1}{c|}{\textbf{66.7}} & \textbf{60.8}     \\ \bottomrule
\end{tabular}%
}
\end{table}


\begin{table}[]
\caption{\textbf{Ablation on Query Point Sampling Strategy.} Success rates (\%) on select MetaWorld tasks when sampling query points from the entire scene versus only from the manipulated object. Scene-level sampling provides critical contextual information, leading to superior performance.}
\label{tab:flow_conditioning_ablation}
\resizebox{\columnwidth}{!}{%
\begin{tabular}{@{}c|ccccc@{}}
\toprule
QT     & Basketball & Box Close & Hammer & Handle Press & Handle Pull \\ \midrule
Object & 93.3       & 51.7      & 48.3   & 86.7         & 38.3        \\
Scene  & \textbf{98.3}    & \textbf{53.3}   & \textbf{68.3} & \textbf{93.3} & \textbf{41.7}        \\ \bottomrule
\end{tabular}%
}
\end{table}

\begin{figure}[]
    \centering
        \begin{subfigure}[t]{\columnwidth}
            \centering
            \includegraphics[width=0.9\textwidth]{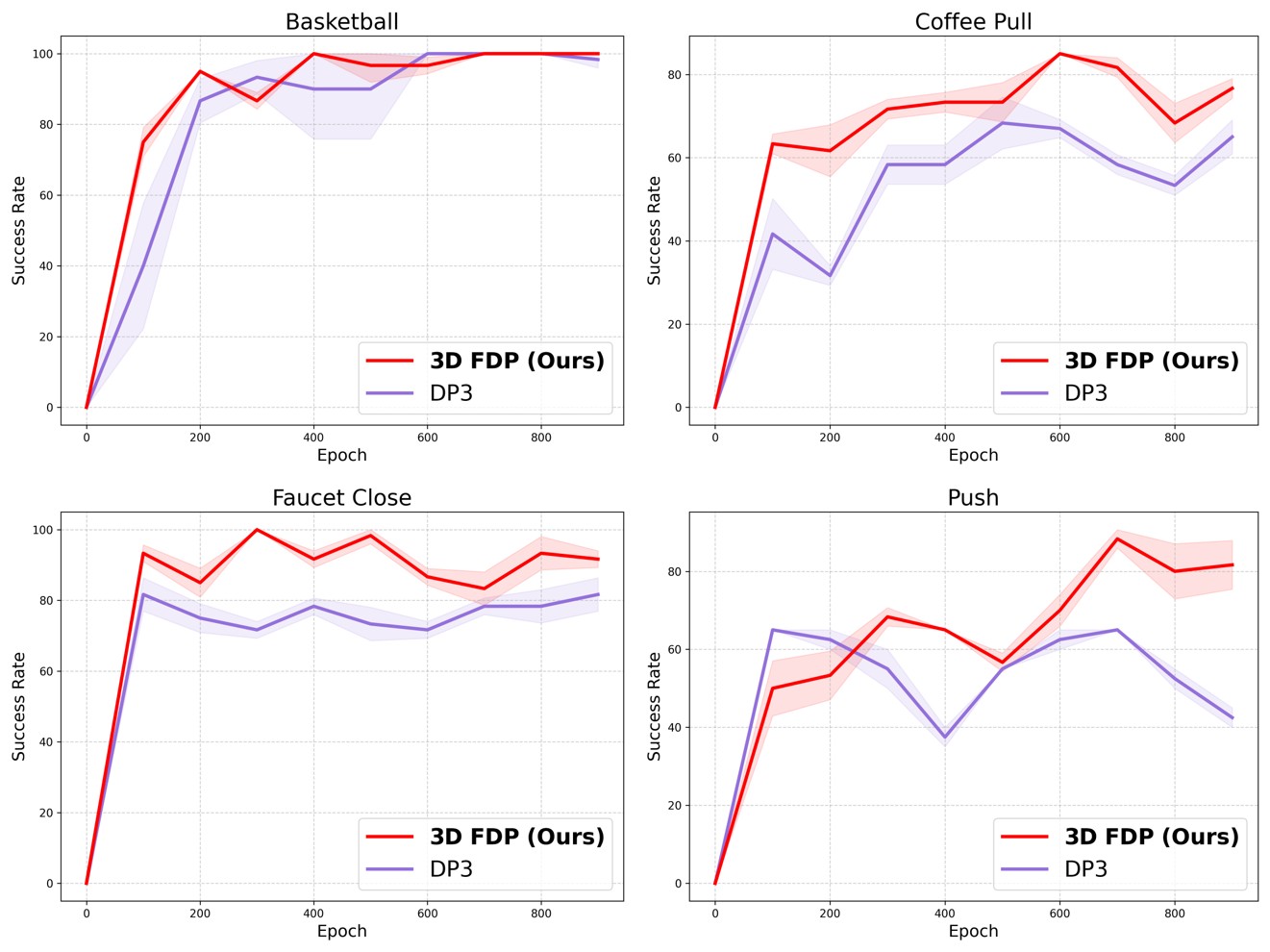}
        \end{subfigure}
        
    \caption{\textbf{Learning efficiency.} 3D FDP converges faster and achieves a higher peak success rate, demonstrating improved sample efficiency and a higher performance ceiling.}
    \label{fig:learning_curve}
\end{figure}

\begin{figure*}[ht!]
    \centering
    \includegraphics[width=0.9\textwidth]{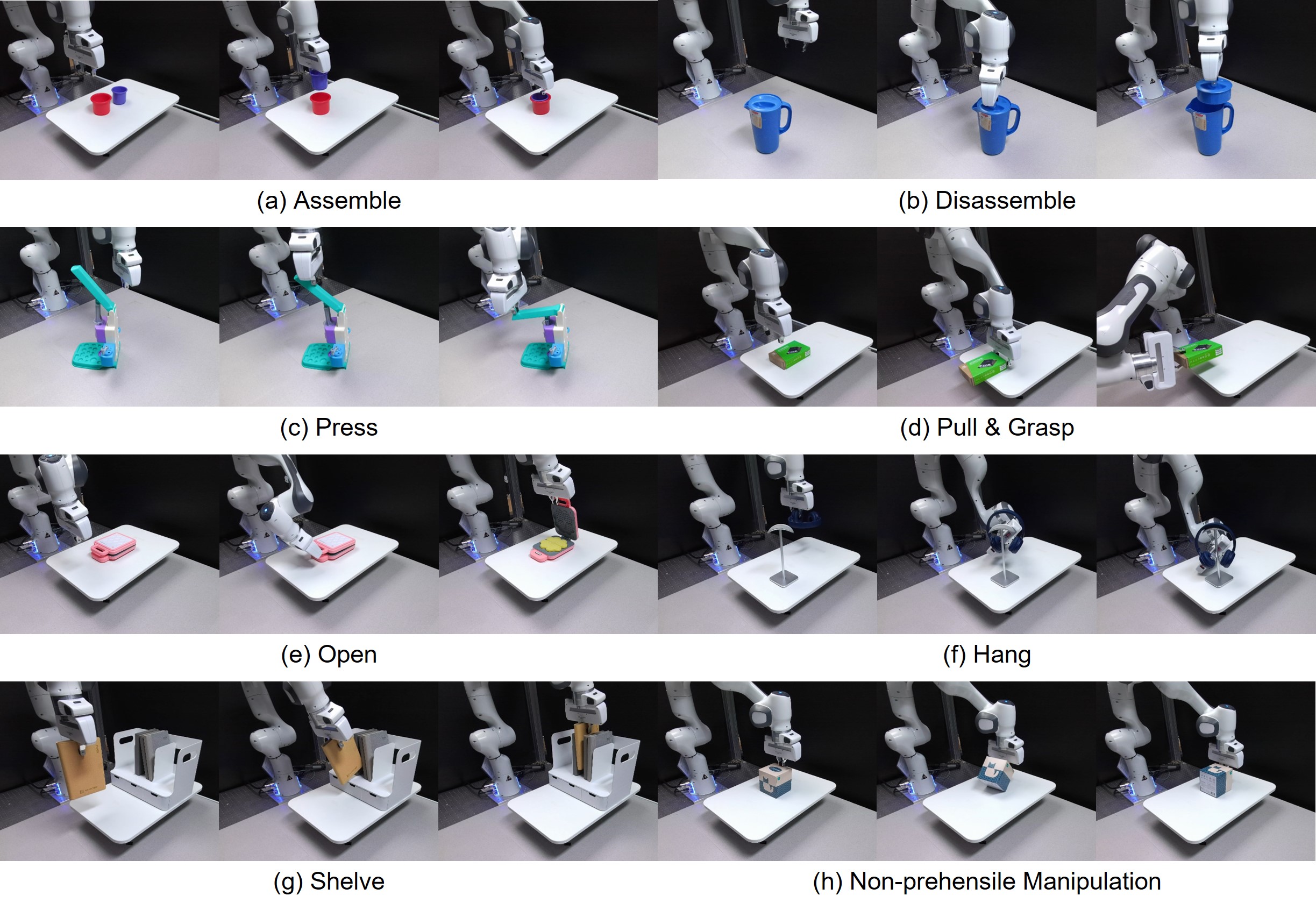}
    \vspace{0.5em}
    
    \resizebox{0.9\textwidth}{!}{%
    \begin{tabular}{@{}cccccccccl@{}}
    \toprule
                                       & \multicolumn{8}{c}{Real-world Tasks}                                                               &      \\ \midrule
    \multicolumn{1}{c|}{Method}        & Assemble & Disassemble & Press & Pull \& Grasp & Open & Hang & Shelve & \multicolumn{1}{c|}{Non-prehensile} & Avg. \\ \midrule
    \multicolumn{1}{c|}{DP3~\cite{ze20243d}}           &  15    &   50    &   25    &     30  &   \textbf{100}  &    0  &     0    & \multicolumn{1}{c|}{0}    &   27.5   \\
    \multicolumn{1}{c|}{3D FDP (Ours)} &  \textbf{60}    &   \textbf{70}    &   \textbf{100}   &     \textbf{40} &    \textbf{100}  &    \textbf{30}  &    \textbf{20}    & \multicolumn{1}{c|}{\textbf{35}}   &   \textbf{56.9}   \\ \bottomrule
    \end{tabular}%
    }
    
    \caption{\textbf{Real-world Tasks and Quantitative Results.} Top: Example tasks used for real-world evaluation. Bottom: Success rates for each method across all tasks.}
    \label{fig:realworld_combined}
\end{figure*}

\noindent\textbf{Simulation Results.}
Table~\ref{tab:main_table} summarizes our results on the MetaWorld benchmark. 3D FDP achieves the highest success rate across all difficulty levels, with particularly notable gains on medium and hard tasks. While DP3 and MBA perform similarly overall, the consistent improvements from 3D FDP suggest that scene-level 3D flow provides a more informative intermediate for action generation. These results indicate that our interaction-aware representation improves policy performance on more complex tasks without compromising effectiveness on simpler ones.

\noindent\textbf{Additional Experiments.}
We conduct further analysis on the MetaWorld hard set, which includes \textit{assembly} (A), \textit{hand-insert} (HI), \textit{pick-out-of-hole} (POH), \textit{pick-place} (PP), \textit{push-back} (PB), and \textit{push} (P), to study the role of flow representations. As shown in Table~\ref{tab:effect_of_nq}, increasing the number of query points from 50 to 200 consistently improves the success rate, indicating that denser flow tracking provides better guidance for action generation.

\begin{figure}[]
    \centering
        \begin{subfigure}[t]{\columnwidth}
            \centering
            \includegraphics[width=0.9\textwidth]{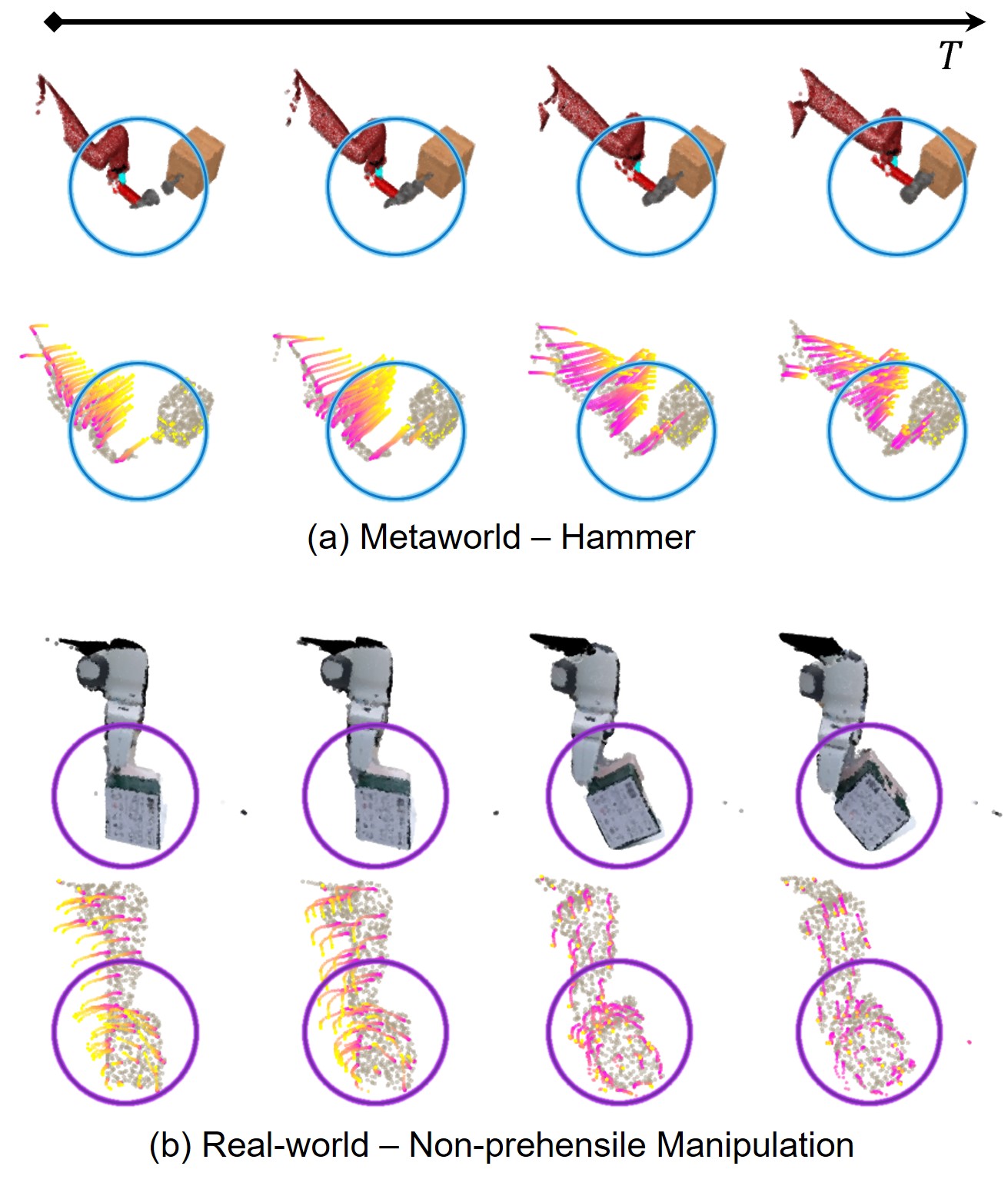}
        \end{subfigure}
        
    \caption{\textbf{Visualization of Predicted 3D Flow.} (a) The 3D FDP predicts flow capturing both the hammer's motion toward the nail and the nail's displacement upon contact. (b) Our model correctly predicts object rotation induced by the robot's linear pushing motion.}
    \label{fig:prediction}
\end{figure}

We also investigate how the spatial distribution of query points affects performance. As shown in Table~\ref{tab:flow_conditioning_ablation}, sampling points from the entire scene leads to higher success rates than restricting them to the manipulated object. This difference becomes more pronounced in tasks like \textit{Hammer} (Fig. \ref{fig:prediction} (a)), where the robot uses a hammer to push a nail into a wall. When query points are limited to the object, the model fails to observe changes in the environment (e.g., nail displacement), which are critical for measuring task progress. In contrast, scene-level sampling captures these interactions more effectively.

Fig.~\ref{fig:learning_curve} presents learning curves for 3D FDP and DP3 on representative tasks including \textit{Basketball}, \textit{Coffee Pull}, \textit{Faucet Close}, and \textit{Push}. While these tasks are not part of the hard set, they involve diverse and challenging interactions. 3D FDP consistently converges faster and achieves higher peak performance, highlighting the efficiency gains from modeling scene-wide flow.

\subsection{Real-world Experiments}

\noindent\textbf{Real Robot Benchmark.}
We further validate our approach on a challenging real-world benchmark of 8 tasks (Fig. \ref{fig:realworld_combined}). Our benchmark is inspired by challenging tasks from prior work, such as \textit{Shelve}~\cite{chen2023predicting, li2023stow}, \textit{Hang}~\cite{you2021omnihang, kuo2024skt} and non-prehensile manipulation~\cite{zhou2023hacman, cho2024corn}, to evaluate policy generalization. Unlike MetaWorld's constrained 4-DoF control, these tasks require a more complex 8-DoF action space (3D position, 4D quaternion, and a binary gripper state), providing a comprehensive evaluation of 3D FDP. The tasks are as follows:
\begin{enumerate}
\item \textbf{Assemble:} Insert an orange cup into a red cup.
\item \textbf{Disassemble:} Remove the lid from a blue bottle.
\item \textbf{Press:} Press a toy button.
\item \textbf{Pull \& Grasp:} Push a box to a graspable position, then grasp it.
\item \textbf{Open:} Open the lid of an articulated waffle maker.
\item \textbf{Hang:} Hang a headset on a stand.
\item \textbf{Shelve:} Insert a book into a cluttered bookshelf.
\item \textbf{Non-prehensile Manipulation:} Rotate a box by applying force to its corner.
\end{enumerate}

\noindent\textbf{Demonstrations and Training.}
We collected 50 expert demonstrations for each task using a GELLO~\cite{wu2024gello} teleoperation device with a Franka Panda manipulator. To introduce slight initial state variations, object poses were randomized within a ±1 cm range during data collection. To maintain consistency with our simulation setup, all policies were trained for 1000 epochs without any data augmentation.

\noindent\textbf{Evaluation Protocol.}
We compare 3D FDP against DP3 as a baseline. Each policy attempts each task 20 times under initial conditions similar to the training data. A trial is considered successful if the final scene state visually matches the goal state achieved in the expert demonstrations, as determined by a human evaluator.

\noindent\textbf{Real-world Results.}
As shown in Table~\ref{fig:realworld_combined}, our real-world experiments reveal a substantial performance gap between 3D FDP and the baseline. Our method achieves an average success rate of 56.9\%, a more than twofold improvement over the 27.5\% from DP3. This result demonstrates the effectiveness of conditioning policies on an explicit model of scene dynamics. Although DP3 shows limited success in structured tasks such as \textit{Assemble} (15\%) and \textit{Press} (25\%), 3D FDP exhibits far greater robustness, achieving success rates of 60\% and 100\% in the same respective tasks. The performance gap widens in scenarios that require long-horizon or complex interactions. For example, in the multi-stage \textit{Pull \& Grasp} task, where the final object pose is variable, 3D FDP's capacity to reason about interaction dynamics over time contributes to its superior performance.

The advantage of our flow-based representation becomes most pronounced in tasks that require precise, contact-rich reasoning, where DP3 consistently fails. DP3 scores 0\% on both \textit{Hang}, which involves generalizing to varied in-hand object poses, and \textit{Shelve}, a difficult task that requires navigating interactions with surrounding objects under heavy occlusion. In these challenging scenarios, 3D FDP succeeds with rates of 30\% and 20\%, respectively. This highlights its ability to model the indirect, scene-level consequences of its actions, a critical factor for manipulation in complex interactive scenarios.

Finally, the \textit{Non-prehensile Manipulation} task clearly illustrates a key limitation of policies that lack an explicit motion model. In this task, a translational push from the robot must induce a rotational motion in the object. 3D FDP effectively manages this complex interaction, achieving a 35\% success rate, while DP3's inability to predict the resulting rotation leads to a 0\% success rate. This indicates that our formulation can capture complex motion relationships that go beyond simple, direct movements, as shown in Fig.~\ref{fig:prediction}~(b). Taken together, these results provide strong evidence that 3D FDP's interaction-aware representation is essential for achieving robust and generalizable performance across a diverse set of real-world manipulation challenges.

\section{Conclusion}

We introduced 3D Flow Diffusion Policy (3D FDP), a visuomotor learning framework that models scene-level 3D flow as an intermediate representation for manipulation. By predicting the future trajectories of sampled query points, 3D FDP captures rich interaction dynamics and uses them to guide action generation through a conditional diffusion process. Our method outperforms existing baselines across 50 tasks in MetaWorld and demonstrates strong generalization to challenging real-world scenarios, including non-prehensile and contact-rich tasks. These results confirm that flow-based representations serve as an effective inductive bias for learning generalizable and robust visuomotor policies.

\section*{Acknowledgment}
\begin{spacing}{0.4}
{\scriptsize This work was supported by the Technology Innovation Program (RS-2024-00442029, Development of Tactile Intelligence in Robotic Hands Based on Tactile Data Learning to Manipulate Irregular Multiple Types of Objects and RS-2024-00423940, Development of Humanoid Robots That Feel Like Humans, Communicate, and Grow through Learning) funded by the Ministry of Trade Industry \& Energy(MOTIE, Korea).}
\end{spacing}

\bibliography{bibtex/bib/references.bib}{}

\begin{thebibliography}{10}
\providecommand{\url}[1]{#1}
\csname url@samestyle\endcsname
\providecommand{\newblock}{\relax}
\providecommand{\bibinfo}[2]{#2}
\providecommand{\BIBentrySTDinterwordspacing}{\spaceskip=0pt\relax}
\providecommand{\BIBentryALTinterwordstretchfactor}{4}
\providecommand{\BIBentryALTinterwordspacing}{\spaceskip=\fontdimen2\font plus
\BIBentryALTinterwordstretchfactor\fontdimen3\font minus \fontdimen4\font\relax}
\providecommand{\BIBforeignlanguage}[2]{{%
\expandafter\ifx\csname l@#1\endcsname\relax
\typeout{** WARNING: IEEEtran.bst: No hyphenation pattern has been}%
\typeout{** loaded for the language `#1'. Using the pattern for}%
\typeout{** the default language instead.}%
\else
\language=\csname l@#1\endcsname
\fi
#2}}
\providecommand{\BIBdecl}{\relax}
\BIBdecl

\bibitem{jang2022bc}
E.~Jang, A.~Irpan, M.~Khansari, D.~Kappler, F.~Ebert, C.~Lynch, S.~Levine, and C.~Finn, ``Bc-z: Zero-shot task generalization with robotic imitation learning,'' in \emph{Conference on Robot Learning}.\hskip 1em plus 0.5em minus 0.4em\relax PMLR, 2022, pp. 991--1002.

\bibitem{cliport2021}
M.~Shridhar, L.~Manuelli, and D.~Fox, ``Cliport: What and where pathways for robotic manipulation,'' in \emph{Conference on robot learning}.\hskip 1em plus 0.5em minus 0.4em\relax PMLR, 2022, pp. 894--906.

\bibitem{brohan2022rt}
A.~Brohan, N.~Brown, J.~Carbajal, Y.~Chebotar, J.~Dabis, C.~Finn, K.~Gopalakrishnan, K.~Hausman, A.~Herzog, J.~Hsu \emph{et~al.}, ``Rt-1: Robotics transformer for real-world control at scale,'' \emph{arXiv preprint arXiv:2212.06817}, 2022.

\bibitem{team2024octo}
O.~M. Team, D.~Ghosh, H.~Walke, K.~Pertsch, K.~Black, O.~Mees, S.~Dasari, J.~Hejna, T.~Kreiman, C.~Xu \emph{et~al.}, ``Octo: An open-source generalist robot policy,'' \emph{arXiv preprint arXiv:2405.12213}, 2024.

\bibitem{kim2024openvla}
M.~J. Kim, K.~Pertsch, S.~Karamcheti, T.~Xiao, A.~Balakrishna, S.~Nair, R.~Rafailov, E.~Foster, G.~Lam, P.~Sanketi \emph{et~al.}, ``Openvla: An open-source vision-language-action model,'' \emph{arXiv preprint arXiv:2406.09246}, 2024.

\bibitem{black2410pi0}
K.~Black, N.~Brown, D.~Driess, A.~Esmail, M.~Equi, C.~Finn, N.~Fusai, L.~Groom, K.~Hausman, B.~Ichter \emph{et~al.}, ``$\pi$0: A vision-language-action flow model for general robot control. corr, abs/2410.24164, 2024. doi: 10.48550,'' \emph{arXiv preprint ARXIV.2410.24164}.

\bibitem{bjorck2025gr00t}
J.~Bjorck, F.~Casta{\~n}eda, N.~Cherniadev, X.~Da, R.~Ding, L.~Fan, Y.~Fang, D.~Fox, F.~Hu, S.~Huang \emph{et~al.}, ``Gr00t n1: An open foundation model for generalist humanoid robots,'' \emph{arXiv preprint arXiv:2503.14734}, 2025.

\bibitem{chi2023diffusion}
C.~Chi, Z.~Xu, S.~Feng, E.~Cousineau, Y.~Du, B.~Burchfiel, R.~Tedrake, and S.~Song, ``Diffusion policy: Visuomotor policy learning via action diffusion,'' \emph{The International Journal of Robotics Research}, p. 02783649241273668, 2023.

\bibitem{ze20243d}
Y.~Ze, G.~Zhang, K.~Zhang, C.~Hu, M.~Wang, and H.~Xu, ``3d diffusion policy: Generalizable visuomotor policy learning via simple 3d representations,'' \emph{arXiv preprint arXiv:2403.03954}, 2024.

\bibitem{ke20243d}
T.-W. Ke, N.~Gkanatsios, and K.~Fragkiadaki, ``3d diffuser actor: Policy diffusion with 3d scene representations,'' \emph{arXiv preprint arXiv:2402.10885}, 2024.

\bibitem{gervet2023act3d}
T.~Gervet, Z.~Xian, N.~Gkanatsios, and K.~Fragkiadaki, ``Act3d: 3d feature field transformers for multi-task robotic manipulation,'' \emph{arXiv preprint arXiv:2306.17817}, 2023.

\bibitem{su2025motion}
Y.~Su, X.~Zhan, H.~Fang, Y.-L. Li, C.~Lu, and L.~Yang, ``Motion before action: Diffusing object motion as manipulation condition,'' \emph{IEEE Robotics and Automation Letters}, 2025.

\bibitem{xian2023chaineddiffuser}
Z.~Xian and N.~Gkanatsios, ``Chaineddiffuser: Unifying trajectory diffusion and keypose prediction for robotic manipulation,'' in \emph{Conference on Robot Learning/Proceedings of Machine Learning Research}.\hskip 1em plus 0.5em minus 0.4em\relax Proceedings of Machine Learning Research, 2023.

\bibitem{black2023zero}
K.~Black, M.~Nakamoto, P.~Atreya, H.~Walke, C.~Finn, A.~Kumar, and S.~Levine, ``Zero-shot robotic manipulation with pretrained image-editing diffusion models,'' \emph{arXiv preprint arXiv:2310.10639}, 2023.

\bibitem{du2023learning}
Y.~Du, S.~Yang, B.~Dai, H.~Dai, O.~Nachum, J.~Tenenbaum, D.~Schuurmans, and P.~Abbeel, ``Learning universal policies via text-guided video generation,'' \emph{Advances in neural information processing systems}, vol.~36, pp. 9156--9172, 2023.

\bibitem{liang2025video}
J.~Liang, P.~Tokmakov, R.~Liu, S.~Sudhakar, P.~Shah, R.~Ambrus, and C.~Vondrick, ``Video generators are robot policies,'' \emph{arXiv preprint arXiv:2508.00795}, 2025.

\bibitem{ni2024generate}
F.~Ni, J.~Hao, S.~Wu, L.~Kou, J.~Liu, Y.~Zheng, B.~Wang, and Y.~Zhuang, ``Generate subgoal images before act: Unlocking the chain-of-thought reasoning in diffusion model for robot manipulation with multimodal prompts,'' in \emph{Proceedings of the IEEE/CVF Conference on Computer Vision and Pattern Recognition}, 2024, pp. 13\,991--14\,000.

\bibitem{luo2024grounding}
Y.~Luo and Y.~Du, ``Grounding video models to actions through goal conditioned exploration,'' \emph{arXiv preprint arXiv:2411.07223}, 2024.

\bibitem{zhao2025cot}
Q.~Zhao, Y.~Lu, M.~J. Kim, Z.~Fu, Z.~Zhang, Y.~Wu, Z.~Li, Q.~Ma, S.~Han, C.~Finn \emph{et~al.}, ``Cot-vla: Visual chain-of-thought reasoning for vision-language-action models,'' in \emph{Proceedings of the Computer Vision and Pattern Recognition Conference}, 2025, pp. 1702--1713.

\bibitem{yang2025gripper}
Y.~Yang, Z.~Cai, Y.~Tian, J.~Zeng, and J.~Pang, ``Gripper keypose and object pointflow as interfaces for bimanual robotic manipulation,'' \emph{arXiv preprint arXiv:2504.17784}, 2025.

\bibitem{metaworld2019}
T.~Yu, D.~Quillen, Z.~He, R.~Julian, K.~Hausman, C.~Finn, and S.~Levine, ``Meta-world: A benchmark and evaluation for multi-task and meta reinforcement learning,'' in \emph{Conference on robot learning}.\hskip 1em plus 0.5em minus 0.4em\relax PMLR, 2020, pp. 1094--1100.

\bibitem{wang2024rise}
C.~Wang, H.~Fang, H.-S. Fang, and C.~Lu, ``Rise: 3d perception makes real-world robot imitation simple and effective,'' in \emph{2024 IEEE/RSJ International Conference on Intelligent Robots and Systems (IROS)}.\hskip 1em plus 0.5em minus 0.4em\relax IEEE, 2024, pp. 2870--2877.

\bibitem{ma2024hierarchical}
X.~Ma, S.~Patidar, I.~Haughton, and S.~James, ``Hierarchical diffusion policy for kinematics-aware multi-task robotic manipulation,'' in \emph{Proceedings of the IEEE/CVF Conference on Computer Vision and Pattern Recognition}, 2024, pp. 18\,081--18\,090.

\bibitem{wang2025hierarchical}
D.~Wang, C.~Liu, F.~Chang, and Y.~Xu, ``Hierarchical diffusion policy: manipulation trajectory generation via contact guidance,'' \emph{IEEE Transactions on Robotics}, 2025.

\bibitem{huang2022flowformer}
Z.~Huang, X.~Shi, C.~Zhang, Q.~Wang, K.~C. Cheung, H.~Qin, J.~Dai, and H.~Li, ``Flowformer: A transformer architecture for optical flow,'' in \emph{European conference on computer vision}.\hskip 1em plus 0.5em minus 0.4em\relax Springer, 2022, pp. 668--685.

\bibitem{shi2023flowformer++}
X.~Shi, Z.~Huang, D.~Li, M.~Zhang, K.~C. Cheung, S.~See, H.~Qin, J.~Dai, and H.~Li, ``Flowformer++: Masked cost volume autoencoding for pretraining optical flow estimation,'' in \emph{Proceedings of the IEEE/CVF conference on computer vision and pattern recognition}, 2023, pp. 1599--1610.

\bibitem{karaev2024cotracker}
N.~Karaev, I.~Rocco, B.~Graham, N.~Neverova, A.~Vedaldi, and C.~Rupprecht, ``Cotracker: It is better to track together,'' in \emph{European conference on computer vision}.\hskip 1em plus 0.5em minus 0.4em\relax Springer, 2024, pp. 18--35.

\bibitem{karaev24cotracker3}
N.~Karaev, I.~Makarov, J.~Wang, N.~Neverova, A.~Vedaldi, and C.~Rupprecht, ``Cotracker3: Simpler and better point tracking by pseudo-labelling real videos,'' in \emph{Proc. {arXiv:2410.11831}}, 2024.

\bibitem{xiao2024spatialtracker}
Y.~Xiao, Q.~Wang, S.~Zhang, N.~Xue, S.~Peng, Y.~Shen, and X.~Zhou, ``Spatialtracker: Tracking any 2d pixels in 3d space,'' in \emph{Proceedings of the IEEE/CVF Conference on Computer Vision and Pattern Recognition}, 2024, pp. 20\,406--20\,417.

\bibitem{bharadhwaj2024track2act}
H.~Bharadhwaj, R.~Mottaghi, A.~Gupta, and S.~Tulsiani, ``Track2act: Predicting point tracks from internet videos enables diverse zero-shot robot manipulation,'' \emph{CoRR}, 2024.

\bibitem{wen2023any}
C.~Wen, X.~Lin, J.~So, K.~Chen, Q.~Dou, Y.~Gao, and P.~Abbeel, ``Any-point trajectory modeling for policy learning,'' \emph{arXiv preprint arXiv:2401.00025}, 2023.

\bibitem{xu2024flow}
M.~Xu, Z.~Xu, Y.~Xu, C.~Chi, G.~Wetzstein, M.~Veloso, and S.~Song, ``Flow as the cross-domain manipulation interface,'' \emph{arXiv preprint arXiv:2407.15208}, 2024.

\bibitem{zhi20253dflowaction}
H.~Zhi, P.~Chen, S.~Zhou, Y.~Dong, Q.~Wu, L.~Han, and M.~Tan, ``3dflowaction: Learning cross-embodiment manipulation from 3d flow world model,'' \emph{arXiv preprint arXiv:2506.06199}, 2025.

\bibitem{yuan2024general}
C.~Yuan, C.~Wen, T.~Zhang, and Y.~Gao, ``General flow as foundation affordance for scalable robot learning,'' \emph{arXiv preprint arXiv:2401.11439}, 2024.

\bibitem{oquab2023dinov2}
M.~Oquab, T.~Darcet, T.~Moutakanni, H.~Vo, M.~Szafraniec, V.~Khalidov, P.~Fernandez, D.~Haziza, F.~Massa, A.~El-Nouby \emph{et~al.}, ``Dinov2: Learning robust visual features without supervision,'' \emph{arXiv preprint arXiv:2304.07193}, 2023.

\bibitem{qi2017pointnet}
C.~R. Qi, H.~Su, K.~Mo, and L.~J. Guibas, ``Pointnet: Deep learning on point sets for 3d classification and segmentation,'' in \emph{Proceedings of the IEEE conference on computer vision and pattern recognition}, 2017, pp. 652--660.

\bibitem{wen2023bundlesdf}
B.~Wen, J.~Tremblay, V.~Blukis, S.~Tyree, T.~M{\"u}ller, A.~Evans, D.~Fox, J.~Kautz, and S.~Birchfield, ``Bundlesdf: Neural 6-dof tracking and 3d reconstruction of unknown objects,'' in \emph{Proceedings of the IEEE/CVF Conference on Computer Vision and Pattern Recognition}, 2023, pp. 606--617.

\bibitem{wen2024foundationpose}
B.~Wen, W.~Yang, J.~Kautz, and S.~Birchfield, ``Foundationpose: Unified 6d pose estimation and tracking of novel objects,'' in \emph{Proceedings of the IEEE/CVF Conference on Computer Vision and Pattern Recognition}, 2024, pp. 17\,868--17\,879.

\bibitem{song2020denoising}
J.~Song, C.~Meng, and S.~Ermon, ``Denoising diffusion implicit models,'' \emph{arXiv preprint arXiv:2010.02502}, 2020.

\bibitem{chen2023predicting}
H.~Chen, Y.~Niu, K.~Hong, S.~Liu, Y.~Wang, Y.~Li, and K.~R. Driggs-Campbell, ``Predicting object interactions with behavior primitives: An application in stowing tasks,'' in \emph{7th Annual Conference on Robot Learning}, 2023.

\bibitem{li2023stow}
Y.~Li, M.~Zhang, M.~Grotz, K.~Mo, and D.~Fox, ``Stow: Discrete-frame segmentation and tracking of unseen objects for warehouse picking robots,'' \emph{arXiv preprint arXiv:2311.02337}, 2023.

\bibitem{you2021omnihang}
Y.~You, L.~Shao, T.~Migimatsu, and J.~Bohg, ``Omnihang: Learning to hang arbitrary objects using contact point correspondences and neural collision estimation,'' in \emph{2021 IEEE International Conference on Robotics and Automation (ICRA)}.\hskip 1em plus 0.5em minus 0.4em\relax IEEE, 2021, pp. 5921--5927.

\bibitem{kuo2024skt}
C.-L. Kuo, Y.-W. Chao, and Y.-T. Chen, ``Skt-hang: Hanging everyday objects via object-agnostic semantic keypoint trajectory generation,'' in \emph{2024 IEEE International Conference on Robotics and Automation (ICRA)}.\hskip 1em plus 0.5em minus 0.4em\relax IEEE, 2024, pp. 15\,433--15\,439.

\bibitem{zhou2023hacman}
W.~Zhou, B.~Jiang, F.~Yang, C.~Paxton, and D.~Held, ``Hacman: Learning hybrid actor-critic maps for 6d non-prehensile manipulation,'' \emph{arXiv preprint arXiv:2305.03942}, 2023.

\bibitem{cho2024corn}
Y.~Cho, J.~Han, Y.~Cho, and B.~Kim, ``Corn: Contact-based object representation for nonprehensile manipulation of general unseen objects,'' \emph{arXiv preprint arXiv:2403.10760}, 2024.

\bibitem{wu2024gello}
P.~Wu, Y.~Shentu, Z.~Yi, X.~Lin, and P.~Abbeel, ``Gello: A general, low-cost, and intuitive teleoperation framework for robot manipulators,'' in \emph{2024 IEEE/RSJ International Conference on Intelligent Robots and Systems (IROS)}.\hskip 1em plus 0.5em minus 0.4em\relax IEEE, 2024, pp. 12\,156--12\,163.

\end{thebibliography}
\bibliographystyle{IEEEtran}

\end{document}